\documentclass{article}
\usepackage{ijcai25}

\usepackage{graphicx}
\usepackage{xcolor}
\usepackage{xspace}
\usepackage{amsmath}
\usepackage{bbm} % For one vector
\usepackage[labelformat=simple]{subcaption}
\usepackage{algorithm}
\usepackage{algpseudocode}
\usepackage{mdwlist}
\usepackage{multirow} % For table
\usepackage{cuted}
\usepackage{afterpage}
\usepackage{stfloats}
\usepackage{hhline}
\usepackage{tabularx}
\usepackage{amsthm}
\usepackage{enumitem}
\usepackage[skip=4pt]{caption}

\usepackage{pifont}
\usepackage{cellspace}
\usepackage{booktabs}
\usepackage{balance}
\usepackage{kotex}

\usepackage{times}
\usepackage{soul}
\usepackage{url}
\usepackage[hidelinks]{hyperref}
\usepackage[utf8]{inputenc}
\usepackage{colortbl}
\usepackage[switch]{lineno}

\definecolor{ForestGreen}{rgb}{0.133, 0.545, 0.133}
\definecolor{torange}{rgb}{0.949, 0.522, 0.}
\newtheorem{problem}{Problem}

\newcolumntype{Y}{>{\centering\arraybackslash}X}
\newtheoremstyle{dotless}{}{}{\itshape}{}{\bfseries}{}{ }{}

\theoremstyle{dotless}

\newcommand{\smallsection}[1]{\vspace{1mm}\noindent\smash{\textbf{#1.}}}
\newcommand{\paper}[1]{\textbf{#1}}
\newcommand{\syndata}{$\{(\mathbf{x}_i, y_i)\}_{i=1}^N$\xspace}

\setlist[itemize]{topsep=1pt, itemsep=0pt}

\let\llncssubparagraph\subparagraph
\let\subparagraph\paragraph
\usepackage{titlesec}
\let\subparagraph\llncssubparagraph
\titlespacing{\section}{0pt}{*1.3}{*1.1}
\titlespacing{\subsection}{0pt}{*1.3}{*1.1}

\graphicspath{ {images/} }

\title{Zero-shot Quantization: A Comprehensive Survey}

\author{
	Minjun Kim$^1$\thanks{Equal Contribution}\and
	Jaehyeon Choi$^1$\footnotemark[1]\and
	Jongkeun Lee$^2$\and
	Wonjin Cho$^2$
	\And
	U Kang$^{1,2}$\thanks{Corresponding Author}
	\affiliations
	$^1$Department of Computer Science and Engineering, Seoul National University \\
	$^2$Interdisciplinary Program in Artificial Intelligence, Seoul National University
	\emails
	\{minjun.kim, jaehyeon\_choi, jklee2, chowonjin0627, ukang\}@snu.ac.kr
}
\begin{document}
	
	\maketitle
	
	\begin{abstract}
		Network quantization has proven to be a powerful approach to reduce the memory and computational demands of deep learning models for deployment on resource-constrained devices.
However, traditional quantization methods often rely on access to training data, which is impractical in many real-world scenarios due to privacy, security, or regulatory constraints.
Zero-shot Quantization (ZSQ) emerges as a promising solution, achieving quantization without requiring any real data.
In this paper, we provide a comprehensive overview of ZSQ methods and their recent advancements.
First, we provide a formal definition of the ZSQ problem and highlight the key challenges.
Then, we categorize the existing ZSQ methods into classes based on data generation strategies, and analyze their motivations, core ideas, and key takeaways.
Lastly, we suggest future research directions to address the remaining limitations and advance the field of ZSQ.
To the best of our knowledge, this paper is the first in-depth survey on ZSQ.
%\blue{Additionally, an up-to-date reading list is maintained at \githublink.} 
	\end{abstract}
	
	\section{Introduction}
	\label{sec:intro}
\begin{figure}[t]
    \centering
%    \vspace{-2mm}
    \includegraphics[width=0.97\linewidth]{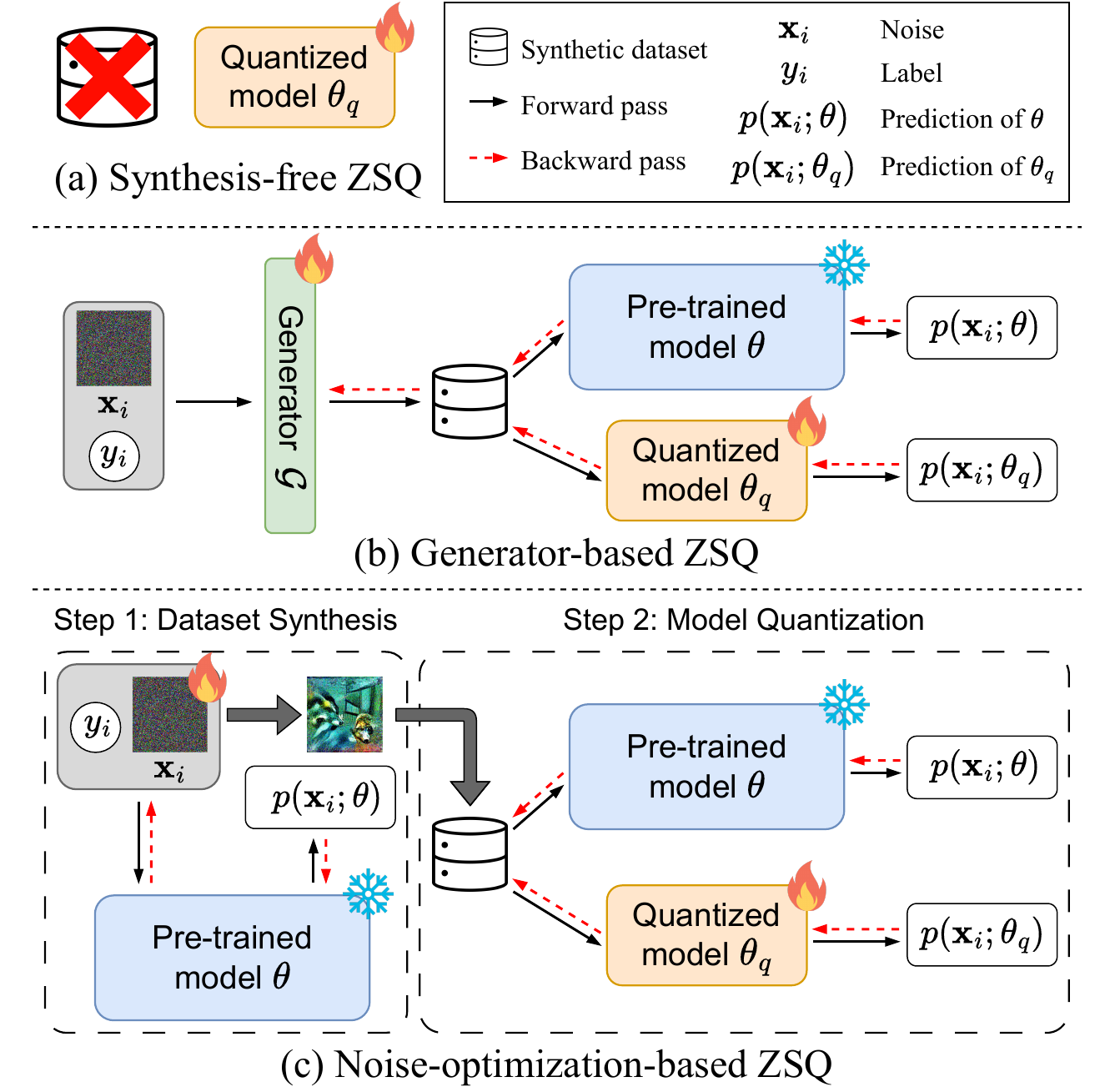}
%    \vspace{1mm}
    \caption{
        Comparison between three categories of Zero-shot Quantization (ZSQ): (a) synthesis-free, (b) generator-based, and (c) noise-optimization-based ZSQ methods.
%		While (a) synthesis-free methods quantize a model without any synthetic data, (b) generator-based and (c) noise-optimization-based methods produce a synthetic dataset \syndata by employing a generator model $\mathcal{G}$ and optimizing randomly initialized noise, respectively.
    }
  \label{fig:overview}
  \vspace{-2mm}
\end{figure}
%%%% Paragraph 1. Model Quantization + Quantization
\textit{How can we accurately quantize a pre-trained model without any data?}
Recent advancements in deep neural networks, including architectures like Convolutional Neural Networks (CNNs)~\cite{ResNet} and Vision Transformers (ViTs)~\cite{ViT}, have achieved the state-of-the-art results in various applications ranging from image classification to visual question answering~\cite{LLaVA}.
However, deploying these models on resource-constrained edge devices remains a significant challenge due to their high memory and computational requirements.
Model compression has emerged as a key technique to address these challenges, offering solutions that reduce model size and computational demands~\cite{GholamiSurvey,Falcon,PruningSurvey}.
Among various compression strategies, network quantization~\cite{BRECQ,SensiMix} stands out by converting high-precision models into a low-bit format, offering high compression and faster inference with minimal performance drop, compared to alternatives such as pruning~\cite{KPrune,PruningSurvey,SPRINT} and knowledge distillation~\cite{PruningAcc,Pea-KD,Towards,PET}. 

%%%% Paragraph 2. Zero-shot Quantization
Zero-shot Quantization (ZSQ)~\cite{DFQ}, also called data-free quantization, addresses a critical limitation in traditional quantization techniques: the dependence on training data.
This is particularly valuable in scenarios where access to original training datasets is restricted due to privacy, security, or regulatory concerns~\cite{GZSQ}.
These limitations are especially pronounced in industries like healthcare, finance, and enterprise services, where sensitive or proprietary data cannot be leveraged for additional model calibration.
Moreover, ZSQ enables the deployment of personalized AI systems, such as mobile AI assistants or security cameras, ensuring efficient performance without compromising data privacy or requiring access to sensitive datasets.
\newpage
  
%%%% Paragraph 3. Limitations of other surveys
Building upon foundational works~\cite{DFQ,KegNet,GDFQ}, numerous studies have further developed the field of ZSQ.
With the rapid growth in research, it has become challenging for researchers to understand the overall trends and essential takeaways of individual studies.
However, existing surveys focus on broader topics
such as general model compression~\cite{SongHanSurvey,DMLabSurvey}
or network quantization~\cite{GholamiSurvey,ContemporarySurvey},
offering only a brief exploration of ZSQ as a subtopic.
This limited coverage restricts researchers from exploring current extent of ZSQ research, distinguishing their key findings, and determining directions for future research.

%%%% Paragraph 4. Contribution of the Survey
In this paper, we conduct a comprehensive and structured survey of ZSQ methods.
We start by formulating the ZSQ problem and exploring three critical challenges.
Next, we summarize existing ZSQ methods (see Table~\ref{tab:overall}) and categorize them based on their data generation strategies: synthesis-free, generator-based, and noise-optimization-based methods.
We illustrate the overall process of each category in Figure~\ref{fig:overview}.
Then, an in-depth examination of these approaches follows, highlighting their motivations, main ideas, and key insights.
Lastly, we propose promising directions for future research, emphasizing unexplored challenges and application scenarios.
Our contributions are summarized as follows:
\begin{itemize}[leftmargin=3.4mm]
	\item \textbf{Overview.}
	We identify major trends in ZSQ algorithms, covering diverse data generation approaches and training scenarios
	(see Figure~\ref{fig:overview} and Table~\ref{tab:overall}).
	\item \textbf{Analysis.}
	We provide a comprehensive review of current ZSQ algorithms, highlighting their motivations, principal ideas, and key findings
	(see Sections~\ref{sec:synthesis_free},~\ref{sec:generator}, and~\ref{sec:noise_optimization}).
	\item \textbf{Discussion.}
	We outline future research directions to advance ZSQ, aiming to guide research toward impactful advancements over current limitations
	(see Section~\ref{sec:further_direction}).
\end{itemize}
%\vspace{2mm} 
	
	\section{Problem Formulation}
	\label{sec:challenge}
	We describe the preliminaries, formulate the ZSQ problem, and discuss the three major challenges that arise in solving it.

\subsection{Preliminaries}
\label{subsec:prelim_quant}

\smallsection{Network Quantization}
Network quantization improves the memory usage and computational efficiency of neural networks by encoding the weight and activations of a given higher-bit network within a lower-bit format.
Quantizing a matrix $\mathbf{W}$ to $B$bit precision involves first rescaling its values to fit within the interval $[-2^{B-1}, 2^{B-1}-1]$.
Each weight is then discretized by mapping to the nearest available integer
% , following the round-to-nearest scheme
~\cite{RTN}.
Given a matrix $\mathbf{W}$, the $B$bit quantized matrix $\mathbf{W}_q$ is calculated as shown in Equation~\ref{eq:eq1}.
\vspace{-1mm}
\begin{equation}
	\label{eq:eq1}
%	\small
	\mathbf{W}_q = \lfloor \frac{\mathbf{W}}{s}+z+\frac{1}{2} \rfloor,
%	~\text{where}~ s=\frac{\beta-\alpha}{2^{B}-1},~
%	z= \lfloor2^{B-1} -\frac{\alpha}{s} +\frac{1}{2}\rfloor,
%\vspace{-1mm}
\end{equation}
where scaling factor $s=(\beta-\alpha)/(2^{B}-1)$,
zero-point $z= -2^{B-1} - {\alpha}/{s}$, and $[\alpha, \beta]$ denotes the clipping range. 
Properly choosing the clipping range $[\alpha, \beta]$ is essential as it defines $s$ and $z$ required for accurate quantization.
A straightforward yet widely adopted approach, known as \textit{min-max quantization}, involves setting $\alpha$ and $\beta$ to the minimum and maximum values of $\mathbf{W}$, respectively.

\smallsection{QAT and PTQ}
Quantization methods are divided into Quantization-aware Training (QAT) and Post-training Quantization (PTQ) based on their training requirements. 
QAT incorporates quantization during fine-tuning, optimizing model performance under quantization constraints while demanding greater computational resources.
In contrast, PTQ is a lightweight approach that quantizes a pre-trained model without additional fine-tuning, making it fast but prone to performance drop.
In ZSQ, while QAT methods rely on min-max quantization, PTQ methods employ advanced techniques such as adaptive rounding~\cite{AdaRound}, block reconstruction~\cite{BRECQ}, and random dropping~\cite{QDrop}, often leading to better performance.

\subsection{Zero-shot Quantization}
\label{subsec:problem_definition}
Given a pre-trained model, Zero-shot Quantization (ZSQ) problem aims to perform network quantization without relying on any real data.
The pre-trained model may address different target tasks, such as image classification, object detection, and semantic segmentation.
We provide the formal definition in Problem~\ref{problem}.

% ================ Problem 1: Zero-shot Quantization =========================

\begin{problem}[Zero-shot Quantization]
	\label{problem}
	% We have a pre-trained model $\theta$ on a task $\mathcal{T}$ and quantization bits $B$.
	We have a model $\theta$ trained on a task $\mathcal{T}$ and quantization bits $B$.
	The goal is to generate a quantized model $\theta_q$ within a $B$bit limit without the use of real data, which shows the best performance on $\mathcal{T}$.
\end{problem}

% ================ END  =========================

\subsection{Main Challenges of ZSQ}
\label{subsec:challenges}

Addressing ZSQ requires overcoming key challenges that arise due to the absence of real data.
This survey highlights how existing approaches tackle
% at least one of
the following challenges:

\begin{itemize}[leftmargin=2.7mm]
	\item \textbf{Knowledge transfer from the pre-trained model.}
	ZSQ relies solely on the information contained in the pre-trained model to recover quantization errors in the absence of real data.
	Therefore, effectively transferring the knowledge, features, or intrinsic characteristics of the pre-trained model to the quantized model is necessary.
	Ensuring this transfer without performance degradation remains a critical issue.
	
	\item \textbf{Discrepancy between real and synthetic datasets.}
	With no access to real datasets, most ZSQ approaches generate synthetic datasets to fine-tune or calibrate the quantized model.
	However, synthetic datasets differ significantly from real-world datasets in various aspects, resulting in severe performance degradation.
	Thus, reducing these disparities to improve the quality of synthetic datasets is crucial. % for accurate quantization.
	
	\item \textbf{Diversity of the problem setting.}
	The scope of the ZSQ problem covers a wide range of tasks, model architectures, experimental conditions, and quantization schemes.
	Hence, it is essential yet challenging to design ZSQ methods that are not only tailored to particular scenarios but also adaptable to diverse tasks or domains.
\end{itemize} 
	
	\section{Categorization}
	\label{sec:categorization}
	% ================ Table 1: Overall Methods =========================

\begin{table*}[ht]
	\caption{A summary of Zero-shot Quantization (ZSQ) methods.
		W\emph{B}A\emph{B} indicates that weights and activations are quantized to \emph{B}bit.
		We compare the ZSQ accuracy [\%] of a ResNet-18 model pre-trained on ImageNet dataset.
		See Section~\ref{sec:categorization} for details.}
	\centering
	\renewcommand{\arraystretch}{0.95}
	\resizebox{\linewidth}{!}{
		\begin{tabular}{l|
				>{\centering\arraybackslash}m{1.8cm}|>{\centering\arraybackslash}m{2cm}|
				>{\centering\arraybackslash}m{2cm}|>{\centering\arraybackslash}m{2cm}|
				>{\centering\arraybackslash}m{1.6cm}|
				>{\centering\arraybackslash}m{1.6cm}>{\centering\arraybackslash}m{1.6cm}}
			\toprule
			\multirow{2}[3]{*}{\textbf{Method} \vspace{1mm}} & \multirow{2}[3]{*}{\textbf{Venue} \vspace{1mm}} & \vspace{1mm} \textbf{Training} & \vspace{1mm} \textbf{Scope of} & \multirow{2}[3]{*}{\textbf{Architecture} \vspace{1mm}} & \multirow{2}[3]{*}{\textbf{\# Images} \vspace{1mm}} & \multicolumn{2}{c}{\textbf{Accuracy (FP = 71.47)}} \\
			% \cmidrule(lr){7-8}
			\cline{7-8}
			& & \textbf{Requirement}\vspace{1mm} & \textbf{Contribution}\vspace{1mm} & & & \vspace{1mm} \textbf{W4A4} & \vspace{1mm} \textbf{W3A3} \\
			
			\midrule
			\rowcolor[rgb]{0.9,0.9,0.9}
			\multicolumn{8}{c}{Synthesis-free ZSQ} \\
			\midrule
			
			DFQ~\shortcite{DFQ} & ICCV & PTQ & Q & CNN & 0 & 55.78 & - \\
			SQuant~\shortcite{SQuant} & ICLR & PTQ & Q & CNN & 0 & 66.14 & 25.74 \\
			UDFC~\shortcite{UDFC} & ICCV & PTQ & Q & CNN & 0 & 63.49 & - \\
			
			\midrule
			\rowcolor[rgb]{0.9,0.9,0.9}
			\multicolumn{8}{c}{Generator-based ZSQ} \\
			\midrule
			
			% DFQ-AKD~\shortcite{DFQ-AKD} & CVPRW & QAT & S, Q & CNN & 102.4K$^*$ & 77.30$^*$ & - \\
			GDFQ~\shortcite{GDFQ} & ECCV & QAT & S, Q & CNN & 1.28M & 60.60 & 20.23 \\
			ZAQ~\shortcite{ZAQ} & CVPR & QAT & S, Q & CNN & 1.28M & 52.64 & - \\
			ARC~\shortcite{ARC} & IJCAI & QAT & S, Q & CNN & 1.28M & 61.32 & 23.37 \\
			Qimera~\shortcite{Qimera} & NeurIPS & QAT & S, Q & CNN & 1.28M & 63.84 & 1.17 \\
			ARC + AIT~\shortcite{AIT} & CVPR & QAT & Q & CNN & 1.28M & 65.73 & - \\
			AdaSG~\shortcite{AdaSG} & AAAI & QAT & S, Q & CNN & 1.28M & 66.50 & 37.04 \\
			AdaDFQ~\shortcite{AdaDFQ} & CVPR & QAT & S, Q & CNN & 1.28M & 66.53 & 38.10 \\
			Causal-DFQ~\shortcite{Casual-DFQ} & ICCV & QAT & S, Q & CNN & 1.28M & 68.11 & - \\
			RIS~\shortcite{RIS} & AAAI & QAT & S & CNN & 1.28M & 67.75 & - \\
			% GenQ (QAT)~\shortcite{GenQ} & ECCV & QAT & S & CNN & 1.2M & 70.03 & 68.18 \\
			GenQ~\shortcite{GenQ} & ECCV & PTQ / QAT & S & CNN & 1K$^\S$ & 69.77$^\S$ & - \\
			\midrule
			\rowcolor[rgb]{0.9,0.9,0.9}
			\multicolumn{8}{c}{Noise-optimization-based ZSQ} \\
			\midrule
			
			DeepInversion~\shortcite{DeepInversion} & CVPR & QAT & S & CNN & 32 & 70.27$^*$ & 64.28$^\dag$ \\
			% GZNQ~\shortcite{GZNQ} & CVPRW & QAT & S, Q & CNN & 1M & 64.50 & - \\
			IntraQ~\shortcite{IntraQ} & CVPR & QAT & S, Q & CNN & 5.12K & 66.47 & 45.51 \\
			HAST~\shortcite{HAST} & CVPR & QAT & S, Q & CNN & 5.12K & 66.91 & 51.15 \\
			% PSAQ-ViT V2~\shortcite{PSAQ-ViTV2} & TNNLS & QAT & S, Q & ViT & 32 & 72.17$^*$ & 68.61$^\dag$ \\
			TexQ~\shortcite{TexQ} & NeurIPS & QAT & S, Q & CNN & 5.12K & 67.73 & 50.28 \\
			PLF~\shortcite{PLF} & CVPR & QAT & Q & CNN & 5.12K & 67.02 & - \\
			SynQ~\shortcite{SynQ} & ICLR & QAT & Q & CNN / ViT & 5.12K & 67.90 & 52.02 \\
			ZeroQ~\shortcite{ZeroQ} & CVPR & PTQ & S, Q & CNN & 1K & 26.04 & - \\
			KW~\shortcite{KW} & CVPR & PTQ & S, Q & CNN & 1K & 69.08 & - \\
			DSG~\shortcite{DSG} & CVPR & PTQ & S & CNN & 1K & 34.53 & - \\
			MixMix~\shortcite{MixMix} & ICCV & PTQ / QAT & S & CNN & 1K$^\S$ & 69.46$^\S$ & - \\
			PSAQ-ViT~\shortcite{PSAQ-ViT} & ECCV & PTQ & S & ViT & 32 & 71.56$^*$ & 65.57$^\dag$ \\
			Genie~\shortcite{Genie} & CVPR & PTQ & S, Q & CNN & 1K & 69.66 & 66.89 \\
			SADAG~\shortcite{SADAG} & ICML & PTQ & S, Q & CNN & 1K & 69.72 & 67.10 \\
			SMI~\shortcite{SMI} & ICML & PTQ & S & ViT & 32 & 70.13$^*$ & 64.04$^\dag$ \\
			CLAMP-ViT~\shortcite{CLAMP-ViT} & ECCV & PTQ & S, Q & ViT & 32 & 72.17$^*$ & 69.93$^\dag$ \\
			\bottomrule
			\addlinespace[1ex]
			\multicolumn{8}{l}{$*$ W8A8 accuracy of DeiT-Tiny~\shortcite{DeiT} model, $\dag$ W4A8 accuracy of DeiT-Tiny~\shortcite{DeiT} model, $\S$ PTQ setting.} \\
		\end{tabular}
	}
	\label{tab:overall}
	 \vspace{-3mm}
\end{table*}

% ================ END  =========================

We categorize existing ZSQ algorithms based on the data generation approach of each algorithm as synthesis-free (see Section~\ref{sec:synthesis_free}), generator-based (see Section~\ref{sec:generator}), and noise-optimization-based (see Section~\ref{sec:noise_optimization}) methods.
Furthermore, for each category, the methodologies are divided into zero-shot QAT and zero-shot PTQ based on their training requirements.

We summarize the key features of ZSQ methods and compare their performance in Table~\ref{tab:overall}.
``Scope of Contribution" identifies whether the main contribution of each method is for data synthesis (S) or network quantization (Q). 
``Architecture" specifies the type of neural network architecture the method is applied to, such as CNN or ViT.
% Experimental settings and performance
Also, we evaluate the 3- and 4-bit ZSQ accuracy of a ResNet-18 model~\cite{ResNet} pre-trained on ImageNet dataset~\cite{ImageNet} for fair benchmarking across methods.
We provide the total number of synthetic samples required to achieve the reported performance as ``\# Images".
For ViT-specific approaches, their ZSQ performance of a DeiT-Tiny model~\cite{DeiT} pre-trained on the ImageNet dataset is reported. 
	
	\section{Synthesis-free ZSQ}
	\label{sec:synthesis_free}
	Synthesis-free ZSQ methods compress a pre-trained model without generating any synthetic data.
These approaches mitigate quantization-induced performance degradation by leveraging structural properties and theoretical foundations of the pre-trained model without generating any synthetic data.

\paper{DFQ}~\cite{DFQ} is a per-tensor weight quantization method that minimizes quantization error through cross-layer equalization and bias correction.
Per-tensor quantization inherently results in higher quantization errors for channels with narrow value ranges when grouped with broader-range channels, as distinct weights are compressed into overly coarse bins.
DFQ addresses this challenge by equalizing the ranges of channel pairs in consecutive layers and adjusting the bias term of each layer based on batch normalization statistics.
DFQ is the first approach to perform weight quantization without any datasets.

\paper{SQuant}~\cite{SQuant} enhances the computational efficiency of Hessian-based quantization by exploiting the structural characteristics of CNNs.
Hessian-based methods effectively optimize quantized models by evaluating the quantization error with the second-order derivative matrix of each weight, but they suffer from substantial computational overhead due to extensive matrix operations.
SQuant improves computational efficiency by performing diagonal Hessian approximation at multiple levels % (element, kernel, and channel levels)
and reduces quantization error by selectively flipping weight signs in decomposed Hessian matrices.
SQuant critically improves the efficiency of Hessian-based quantization without data dependency.

\paper{UDFC}~\cite{UDFC} proposes a hybrid model compression algorithm that integrates pruning and quantization techniques.
UDFC addresses the damage resulting from pruning or quantization by leveraging the weighted combination of remaining undamaged channels.
Specifically, after pruning or quantizing the $l$th layer, UDFC adjusts the $(l+1)$th layer through a theoretically derived closed-form solution to reduce reconstruction error.
UDFC introduces the first unified approach for combining both quantization and pruning in a zero-shot compression setting.

	\section{Generator-based ZSQ}
	\label{sec:generator}
	Generator-based ZSQ methods employ an independent generator model $\mathcal{G}$ to produce synthetic datasets for quantizing pre-trained models.
Specifically, they generally train a Generative Adversarial Network (GAN)-based generator~\cite{GAN} such as DCGAN and ACGAN from scratch.
An ideal generator $\mathcal{G}$ would generate a synthetic dataset \syndata with a distribution that closely resembles a real dataset.
Therefore, generator-based algorithms aim to train a generator $\mathcal{G}$ to produce synthetic datasets that closely match real data distributions by incorporating crucial information extracted from pre-trained models.
One example of such information that is commonly important for quantizing CNN models is Batch Normalization Statistics (BNS).
BNS refers to the mean and variance calculated during CNN training, which reflects the distribution of the training data processed by each batch normalization layer.
Many studies reduce the L2 norm between the BNS of real and synthetic datasets across all $L$ layers, enforcing a regularization constraint on the synthetic dataset during updates.

\subsection{Generator-based Zero-shot QAT}
\label{subsec:generator_QAT}

Most generator-based algorithms generate images at each step to refine a generator model to mimic real data accurately.
Instead of fine-tuning the quantized model only with the final images from the trained generator, QAT methods exploit images produced at every training stage of the generator as they inherently retain the pre-trained model’s knowledge.
Specifically, these approaches involve alternating or adversarial learning to balance quantization efficiency and output quality.
Generator-based QAT methods share an identical experimental setup with 1.28M synthetic images produced over 400 epochs, each consisting of 200 batches of 16 samples.
%

% \paper{DFQ-AKD}~\cite{DFQ-AKD}

\paper{GDFQ}~\cite{GDFQ} is the first ZSQ method that incorporates a generator to produce a synthetic dataset.
GDFQ refines the generator and the quantized model through alternating optimization.
The generator aims to produce synthetic data that allows the pre-trained model to predict labels accurately, while the quantized model learns to classify these synthetic images correctly to minimize the performance gap.
This strategy progressively improves both data quality and model performance, achieving robust performance.

\paper{ZAQ}~\cite{ZAQ} employs adversarial learning to optimize quantization by introducing competition between a generator and a quantized model.
Adversarial learning follows a minimax optimization, where the generator maximizes the performance gap between the pre-trained and quantized models while the quantized model refines itself to minimize it.
Additionally, ZAQ incorporates activation regularization to guide the generator and feature-level knowledge distillation to better encourage the quantized model to mimic the pre-trained model.
This adversarial learning framework serves as a basis for generator-based ZSQ research, accelerating the advancement of optimization algorithms in this domain.
	
\paper{ARC}~\cite{ARC} or AutoReCon automatically determines a generator architecture by leveraging neural architecture search.
Most existing techniques employ GAN-based architectures tailored for image generation instead of model compression.
To resolve this issue, ARC employs neural architecture search to find generator architectures suited for compression.
With an optimal generator, ARC outperforms existing methods in 3-bit quantization by up to 11\%p.

\paper{Qimera}~\cite{Qimera} trains a generator to produce boundary supporting samples, aiming to improve quantization performance.
Boundary supporting samples are located near the decision boundaries of the pre-trained model; however, existing methods lack such samples in synthetic datasets, which limits the ability of the quantized model to learn the decision boundaries of the pre-trained model effectively.
To address this limitation, Qimera encourages the generator to synthesize samples near the decision boundaries by utilizing superposed latent embeddings.
Qimera, when combined with existing ZSQ methods, improves up to 9\%p in quantization performance under 4-bit quantization settings.

\paper{AIT}~\cite{AIT} emphasizes that Cross-Entropy (CE) loss hinders the optimization process when training quantized models with synthetic datasets.
While previous approaches employ both CE and Kullback–Leibler divergence (KL) losses to optimize quantized models, the authors observe two key points: 1) the conflict between CE and KL losses, and 2) the superior generalizability of KL loss.
Motivated by these observations, AIT eliminates the CE loss and focuses exclusively on KL loss to optimize the quantized model.
Additionally, it manipulates gradients to ensure that a minimum ratio of integer values in the quantized model is updated during each optimization step, thereby enhancing optimization efficiency.
AIT integrates easily with other generator-based ZSQ methods to improve performance and promote optimization efficiency.

\paper{AdaSG}~\cite{AdaSG} measures `sample adaptability,' which is the contribution of synthetic images in training quantized models, and proposes a novel optimization algorithm for this metric.
Existing ZSQ methods fail to fully recover performance degradation caused by quantization since they neglect the characteristics of quantized models during synthetic image generation.
AdaSG incorporates sample adaptability by reformulating the ZSQ problem into a zero-sum game between the generator and the quantized model.
AdaSG presents the first game-theoretical formulation of the ZSQ problem, establishing a novel problem-solving paradigm based on sample adaptability.

\paper{AdaDFQ}~\cite{AdaDFQ} proposes a boundary-based optimization method that achieves stable optimization through effectively controlling sample adaptability.
Existing ZSQ methods that incorporate sample adaptability encounter optimization instability, leading to either overfitting or underfitting issues.
AdaDFQ defines two boundaries determined by agreement and disagreement based on the predictions of pre-trained and quantized models, and optimizes the margin between these boundaries to ensure that generated samples maintain adaptability with respect to the quantized model.
AdaDFQ enhances the stability of the optimization process in ZSQ while maintaining sample adaptability through its boundary-based optimization approach.

\paper{Causal-DFQ}~\cite{Casual-DFQ} introduces causal reasoning to disentangle content and style for improving synthetic dataset quality.
Content captures task-relevant features, while style represents irrelevant attributes that do not influence model decisions.
Unlike existing methods that rely on only statistical information (e.g., BNS), Causal-DFQ models these factors separately by constructing a causal graph.
It designs a content-style-decoupled generator to synthesize images by independently modulating content and style.
This is the first approach to introduce causal relationships in ZSQ, enhancing the diversity and robustness of synthetic data.
%
%\paper{Causal-DFQ}~\cite{Casual-DFQ} introduces causal reasoning to extract relevant factors from a pre-trained model for a training generator.
%Existing methods extract only statistical information (e.g., BNS) from a pre-trained model for synthetic data generation.
%To overcome this limitation, Causal-DFQ constructs a causal graph based on content as the relevant factor and style as the irrelevant factor, and designs a content-style-decoupled generator that generates synthetic images by taking these two independent factors as input.
%This study is the first to introduce causal relationships in the ZSQ setting and effectively ensures the diversity of synthetic images by modulating style factors.

\paper{RIS}~\cite{RIS} encourages the generator to produce diverse synthetic images that contain semantic information.
The authors observe that synthetic images are more vulnerable to perturbations compared to real images, indicating a lack of semantic information.
RIS explicitly models robustness against perturbations at both feature and prediction levels by applying perturbations to synthetic images and weights of pre-trained models. 
Furthermore, RIS employs soft labels instead of hard labels as input to the generator to facilitate the creation of diverse synthetic images.
RIS demonstrates that training generators to produce perturbation-robust synthetic images improves the performance of quantized models. 

\subsection{Generator-based Zero-shot PTQ}
\label{subsec:generator_PTQ}

Some generator-based ZSQ methods adopt PTQ scheme by generating synthetic datasets using a pre-trained generator, such as diffusion model, as this approach does not require a training process. 
Consequently, they adopt PTQ scheme effectively leveraging characteristics of the synthetic datasets and the architecture of the pre-trained model.

% 종근
\paper{GenQ}~\cite{GenQ} introduces a novel approach to synthesizing reliable data using diffusion-based text-to-image models.
Existing methods struggle to generate semantically rich, high-resolution data due to the complexity of mapping low-dimensional labels to high-dimensional images, resulting in distribution gaps compared to real data.
GenQ addresses this challenge by generating a synthetic dataset with Stable Diffusion, introducing three filtering techniques to minimize distribution gaps and improve the quality of synthetic data:
1) energy score filtering, which identifies in-distribution data by measuring the confidence of model predictions through energy scores, 2) BNS distribution filtering, which aligns the activation statistics of synthetic data with those of real data, and 3) patch similarity filtering, which ensures diversity in visual representations for ViTs.
GenQ pioneers using text-to-image diffusion models to generate synthetic data for ZSQ, maintaining high quantization accuracy across varied settings.

% \paper{ZSQ-MPQ}~\cite{ZSQ-MPQ} 
	
	\section{Noise-optimization-based ZSQ}
	\label{sec:noise_optimization}
	Noise-optimization-based ZSQ algorithms directly optimize noise to generate the synthetic dataset by iteratively updating the input itself rather than the parameters of a generator model.
They universally follow a two-step scheme by first optimizing randomly initialized noise to generate a synthetic dataset \syndata with size $N$ (step 1: dataset synthesis) and then quantizing the pre-trained model with those samples (step 2: model quantization).
Similar to generator-based methods, BNS loss serves as the baseline loss, aligning the statistics of all $L$ batch normalization layers between synthetic and real datasets.
Additionally, Inception Loss (IL) is optimized to reduce the cross-entropy between sample labels $\{y_i\}_{i=1}^N$ and predictions $\{p(\mathbf{x}_i; \theta)\}_{i=1}^N$ of the pre-trained model $\theta$, encouraging the model to incorporate class-specific details.
The main focus is to produce a synthetic dataset that mimics the training dataset of the pre-trained model; each algorithm identifies and mitigates the key discrepancies between real and synthetic datasets.

\subsection{Noise-optimization-based Zero-shot QAT}
\label{subsec:noise_QAT}
Following dataset synthesis, QAT methods first simply quantize the pre-trained model following min-max quantization and then fine-tune it with the synthetic dataset.
These methods share a common experimental setting generating a total of 5,120 images, given that a larger synthetic dataset results in higher model accuracy.

\paper{DeepInversion}~\cite{DeepInversion} pioneers the generation of synthetic datasets for knowledge transfer in scenarios without prior data.
The method has two objectives: enforcing feature similarity at all layers between real and synthetic datasets with BNS loss, and generating samples that challenge the quantized model but align with the pre-trained model by penalizing output distribution similarities between the models through training competition.
DeepInversion is a general framework that generates a synthetic dataset to transfer knowledge from pre-trained models, applicable not only to ZSQ but also to various data-free tasks such as pruning, knowledge distillation, and continual learning.

% \paper{GZNQ}~\cite{GZNQ}

\paper{IntraQ}~\cite{IntraQ} identifies intra-class heterogeneity as a key factor for improving synthetic dataset quality.
Most approaches generate consistent samples within each class because they prioritize overall dataset distribution and inter-class separation.
IntraQ promotes intra-class diversity by varying object scale and location with local object reinforcement, distributing class features broadly across images with a marginal distance constraint, and mitigating overfitting with soft IL.
Their findings stand out through their high performance, advancing beyond traditional methods which face challenges with bit assignments under 4bit, and taking the first step into extreme low-bit ZSQ at 3bit.

% 종근
% \paper{IntraQ}~\cite{IntraQ} retains intra-class heterogeneity in synthetic images to improve the performance of ZSQ models.
% Existing methods generate homogeneous synthetic images by focusing on overall dataset distribution or class-wise separation, leading to poor generalization of synthetic images where intra-class variability is critical.
% IntraQ addresses this issue by introducing three strategies: 1) local object reinforcement to place target objects at different scales and locations, 2) a marginal distance constraint to distribute class-relevant features broadly across the image, and 3) soft inception loss to mitigate overfitting by replacing one-hot labels with soft labels, fostering richer and more diverse feature representations.
% IntraQ preserves intra-class heterogeneity in synthetic data, moving beyond traditional methods focused solely on inter-class separation.

\paper{HAST}~\cite{HAST} highlights that including hard samples in a synthetic dataset enhances ZSQ performance.
Previous methods perform poorly on hard images,
where the difficulty of an image~\cite{GHM} indicates how likely a model is to misclassify it, because their synthetic datasets lack challenging samples.
Therefore, HAST increases the portion of difficult samples within the synthetic dataset for both pre-trained and quantized models by optimizing hard-sample-enhanced IL and promoting sample difficulty, respectively.
HAST shows outstanding performance, comparable to that of fine-tuning with real datasets.

% 원진
% \paper{HAST}~\cite{HAST} reduces the discrepancy of the difficulty distribution between the synthetic and real datasets, where the difficulty is the probability of incorrectly classifying a dataset.
% Previous methods experienced performance degradation in the difficult test dataset as these methods train only with an easy synthetic dataset.
% HAST addresses this by optimizing cross-entropy loss and applying a perturbation to generate difficult synthetic datasets that suit pre-trained and quantized models.
% It also optimizes feature alignment loss to distill knowledge well.
% HAST performs better than the quantized model fine-tuned with real datasets in some settings, showing the need for difficult training datasets.

% \paper{PSAQ-ViT V2}~\cite{PSAQ-ViTV2} \TODO{.}

\paper{TexQ}~\cite{TexQ} focuses on minimizing discrepancies in the distributions of texture features.
Texture describes the spatial arrangement of pixel intensities forming visual patterns, vital for CNNs which classify images based on surface characteristics; prior methods often underperform due to insufficient texture features in the synthetic dataset.
TexQ resolves this issue by directing samples to accurate per-class calibration centers, ensuring texture alignment with LAWS energy loss, and containing texture features in shallow layers with layered BNS loss.
The authors empirically validate that adding sufficient texture features enhances inter-class distance, leading to better ZSQ performance.

% 원진
% \paper{TexQ}~\cite{TexQ} improves the quality of a synthetic dataset by reducing the texture feature deviation between real and synthetic datasets.
% This is crucial as CNNs primarily classify images with the surface characteristics of objects, known as textures.
% However, prior methods often perform poorly due to insufficient texture features in a synthetic dataset.
% TexQ solves this by
% aligning the proportion of texture features by applying texture filters, aligning other important features through BNS loss, and diversifying the synthetic dataset by applying mixup knowledge distillation.
% TexQ is the first approach to generate the synthetic dataset with unique feature alignment frameworks.

\paper{PLF}~\cite{PLF} is the first approach to evaluate the synthetic dataset before quantization.
After synthesizing the dataset, PLF separates it into high- and low-reliable groups based on self-entropy computed from the probabilities of a pre-trained model, where lower self-entropy indicates more confident predictions.
Then, PLF assigns the second highest probability label as an auxiliary label for low-reliable data to soften supervised learning and reduce the risk from misleading labels.
Unlike methods such as HAST, which determine confidence based on image difficulty~\cite{GHM}, PLF adopts self-entropy to measure confidence effectively.

\paper{SynQ}~\cite{SynQ} emphasizes three major limitations when fine-tuning with synthetic datasets: mismatches of amplitude distribution in the frequency domain, predictions based on off-target patterns, and harmful effects of erroneous hard labels for hard samples.
SynQ addresses these challenges by introducing a low-pass filter to reduce high-frequency noise, aligning class activation maps to identify correct image regions, and excluding cross-entropy loss for hard samples to reduce ambiguity.
SynQ is the first work to explore both CNNs and ViTs, showing high adaptability toward various models and dataset synthesis algorithms.

\subsection{Noise-optimization-based Zero-shot PTQ}
\label{subsec:noise_PTQ}
PTQ algorithms adjust the scaling factor $s$ and zero-point $z$ or apply advanced quantization techniques instead of directly updating model parameters.
Compared to QAT, these methods require smaller synthetic datasets for calibration, typically evaluated with sets of 1,000 for CNNs and 32 for ViTs.
	
\paper{ZeroQ}~\cite{ZeroQ} pioneers zero-shot PTQ as the first method of its kind.
ZeroQ first generates a synthetic dataset by optimizing a set of random noise with BNS loss and then employs it to determine the clipping range $[\alpha, \beta]$.
This approach further extends to mixed-precision quantization by assigning bit precision configurations based on a Pareto frontier.
ZeroQ guides future researches toward enhancing synthetic datasets while maintaining its sample-driven approach for clipping range selection.

\paper{KW}~\cite{KW} proposes generating class-specific synthetic datasets by integrating BNS loss and IL.
In contrast to previous techniques which do not produce class-specific data, this approach generates images corresponding to each image class.
IL aligns the classifier probabilities of images with their pre-assigned labels, thereby injecting desired class information into each image.
Combining BNS loss and IL achieves high performance in 4bit quantization, showing only a 2-3\%p accuracy drop compared to the original model, making it a baseline loss for future studies.

\paper{DSG}~\cite{DSG} addresses the homogenization issues in synthetic datasets limited by BNS at distribution and sample levels.
The authors report that feature distributions in synthetic datasets overfit to BNS (distribution-level homogenization), and identical optimization objectives result in excessive sample similarity (sample-level homogenization).
DSG improves data diversity by slack distribution alignment, which relaxes BNS constraints, and layer-wise sample enhancement, which reinforces per-sample loss.
DSG demonstrates that addressing the homogenization caused by BNS significantly reduces overfitting and boosts the diversity of synthetic datasets.

\paper{MixMix}~\cite{MixMix} focuses on reducing biases at feature and label levels in generated samples.
The authors observe that features are biased as they originate from a specific model, preventing direct application across different architectures.
Additionally, labels of generated samples are biased due to inexact inversion from low-dimensional to high-dimensional data.
MixMix introduces two mixing techniques: feature mixing to construct a universal feature space across models and data mixing to ensure accurate label representation by combining synthetic samples and labels.
The results indicate that synthetic datasets are inherently biased and suitable only for their respective pre-trained models.
	
\paper{PSAQ-ViT}~\cite{PSAQ-ViT} is the first ZSQ method tailored for ViTs.
CNN-based methods rely mainly on BNS loss, making them unsuitable for ViTs with layer normalization and self-attention modules.
PSAQ-ViT optimizes the Patch Similarity Entropy (PSE) loss, aiming to maximize the diversity of cosine similarities between the outputs of self-attention layers for two different input patches, thereby generating samples that reflect self-attention diversity.
PSE loss serves as the baseline loss in ViT-based ZSQ, similar to BNS loss in CNN-based ZSQ.
%The authors further extend the paper into a QAT setting by employing a cyclic evolution between the generated samples and the quantized model in a competitive manner~\cite{PSAQ-ViTV2}.

\paper{Genie}~\cite{Genie} integrates generator-based methods with noise-optimization-based algorithms.
Genie first trains a generator to extract the common knowledge of the input domain, then indirectly optimizes the synthetic dataset by learning from the trained generator.
They distill the knowledge with swing convolution to avoid information loss while maintaining computational efficiency.
Genie further adopts advanced PTQ techniques such as adaptive rounding~\cite{AdaRound}, block-wise reconstruction~\cite{BRECQ}, and random dropping~\cite{QDrop}, leading to improved performance. 
%compared to simply adjusting the clipping range.

\paper{SADAG}~\cite{SADAG} proposes a sharpness-aware generation method to enhance the generalization of the quantized model.
Optimizing the noise with sharpness-aware minimization reduces both loss value and sharpness, leading the image to a flat local optimum.
Therefore, SADAG creates samples with less loss sharpness by applying gradient matching.
The authors find that optimizing for loss sharpness generates smoother images with reduced color variation while maintaining semantics, leading to better performance.

\paper{SMI}~\cite{SMI} generates only the essential parts of an image to accelerate generation speed while maintaining ZSQ performance.
Existing methods rely on dense model inversion to produce all pixels of a fixed-size image, unnecessarily spending equal time on unimportant backgrounds and often introducing label errors by generating objects from other classes.
SMI addresses this issue by evaluating patch-level importance through ViT attention scores and stopping optimization for irrelevant patches, resulting in a sparse dataset.
SMI achieves up to 3.79$\times$ faster image generation with 77\% sparsity while achieving similar or up to 0.78\%p higher accuracy in W4A8 quantization, highlighting the effectiveness of sparse model inversion.

% 원진
% \paper{SMI}~\cite{SMI} introduces the first method to generate sparse images, images only with important sections.
% This accelerates the generation process and removes unnecessary information in images.
% Previous methods generate the entire image area, which includes unintended objects or noisy backgrounds, resulting in inferior performance.
% SMI generates sparse images in two key steps. They first identify the relevance of semantic patches by calculating attention scores. Secondly, they apply progressive stopping to remove irrelevant semantic patches in model inversion.
% SMI is the first approach to address the efficiency of sparse images, which reduces memory usage and computation while maintaining accuracy.

\paper{CLAMP-ViT}~\cite{CLAMP-ViT} adopts contrastive learning to integrate semantic information into the synthetic dataset.
The authors point out that patch similarity in PSAQ-ViT assumes equal importance for all patches without considering spatial sensitivity, failing to capture semantically meaningful inter-patch relations. 
To overcome this problem, CLAMP-ViT introduces a patch-level contrastive learning framework designed to enhance the semantic quality of synthetic data tailored for ViT models.
Furthermore, CLAMP-ViT supports mixed-precision quantization through layer-wise evolutionary search, which determines the optimal bit-width and quantization parameters.

% 종근
% \paper{CLAMP-ViT}~\cite{CLAMP-ViT} leverages patch-based contrastive learning to generate semantically meaningful synthetic data and employs layer-wise evolutionary search to maximize performance.
% Existing studies fail to effectively utilize semantic relationships between patches, generating simplistic and semantically vague data that lead to suboptimal quantization accuracy.
% CLAMP-ViT proposes patch-level contrastive learning to generate semantically rich synthetic data and layer-wise evolutionary search to mitigate the challenges posed by non-smooth loss landscapes while identifying optimal quantization parameters.
% CLAMP-ViT enables both fixed- and mixed-precision quantization without requiring real data, achieving up to 3 percent improvement in accuracy across diverse vision tasks.

\vspace{1mm} 
	
	\section{Further Research Directions}
	\label{sec:further_direction}
	
ZSQ is a rapidly growing field with substantial potential.
Although notable advances have been made in ZSQ, numerous research directions remain open for future exploration.

\begin{itemize}[leftmargin=3.2mm, topsep=2mm]
    \item \textbf{More principled analysis on synthetic datasets.}
    As mentioned in Section~\ref{subsec:challenges}, reducing the discrepancy between real and synthetic datasets is one of the significant challenges for ZSQ methods.
    However, researchers often focus on individual features (e.g., intra-class heterogeneity, amplitude distribution, etc.) instead of investigating the underlying causes of their differences.
    Conducting a deeper analysis of synthetic datasets might fundamentally enhance ZSQ methods, going beyond mere patchwork solutions.
	
    \item \textbf{Broader application to various tasks and domains.}
    While ZSQ encompasses a wide range of settings, most research concentrates on the image domain, focusing primarily on CNN and ViT models.
    Specifically, they set task $\mathcal{T}$ mainly as image classification, with a few work on object detection~\cite{PSAQ-ViT,Casual-DFQ} or semantic segmentation~\cite{DFQ}.
    Extending ZSQ research to various tasks such as language, multi-variate, or graph-based ones is crucial for advancing the field~\cite{Mustard,AugWard}.

    \item \textbf{Theoretical exploration of ZSQ.}
    A formal investigation into the theoretical limits of zero-shot quantization performance is essential for understanding its full potential.
    This includes defining the upper bounds on model accuracy without access to real data, and exploring how quantization bit-widths affect performance degradation.
    Furthermore, identifying the mathematical principles underlying ZSQ is crucial for developing more robust algorithms.

    \item \textbf{Faster generation of synthetic datasets.}
    Increasing the size of synthetic datasets enhances the performance of quantized models by providing more diverse training data~\cite{SynQ}.
    However, existing algorithms require a significant amount of time, requiring 1 to 4 RTX 4090 GPU hours for generating 5,120 images with a resolution of 224 $\times$ 224.
    %To apply these methods practically, researchers should focus on optimizing performance within equal time frames instead of comparing based on identical image quantities.
    Therefore, reducing the time required for data generation is a vital topic for future exploration. 

    \item \textbf{Combining other model compression techniques.}
    Current ZSQ methods achieve competitive results in 4bit quantization but struggle to retain performance in 3bit or lower-bit quantization.
    Integrating quantization with other model compression methods such as pruning, weight sharing, or low-rank approximation might be a key to achieve higher compression rate while maintaining accuracy.
    % \TODO{third sentence if needed}.

    \item \textbf{Evaluating practical impact on real-world scenarios.}
    The importance of ZSQ lies in its applications for handling real-world scenarios with limited data.
    However, current ZSQ methods present experimental results solely on benchmark datasets and models.
%    , leaving their real-world applicability underexplored.
    The lack of evaluation restricts researchers from validating the practicality of proposed methods, limiting the reliability of ZSQ approaches.
    Hence, evaluating ZSQ methods under practical settings is required to compare their real-world applicability. 

    \item \textbf{Diverse problem settings.}
    Recent advancements in ZSQ have led to the exploration of diverse problem settings beyond conventional scenarios.
    These include setups such as few-instance quantization, where 1 to 10 real images are available, or cases leveraging pre-trained diffusion models~\cite{GenQ}.
    Extending ZSQ to real-time quantization and edge-device deployments are vital future directions in enhancing the practical utility.

\end{itemize}

% \vspace{1mm} 
	
	\section{Conclusion}
	\label{sec:conclusion}
	In this paper, we perform an extensive survey of ZSQ.
%To the best of our knowledge, this paper provides a comprehensive overview of ZSQ methods for the first time.
We first present the preliminaries and formulate the problem with its key challenges. 
Then, we review ZSQ algorithms with a novel taxonomy and categorize them based on data generation approaches and training requirements.
Specifically, we detail the motivations, ideas, and findings of each paper comprehensively.
Finally, we introduce promising research topics on ZSQ, providing insights to resolve current constraints. 
	
	\section*{Acknowledgments}
	This work was supported by Institute of Information \& communications Technology Planning \& Evaluation (IITP) grant funded by the Korea government (MSIT)
	[No. RS-2020-II200894, Flexible and Efficient Model Compression Method for Various Applications and Environments],
	[No. RS-2021-II211343, Artificial Intelligence Graduate School Program (Seoul National University)],
	and [No. RS-2021-II212068, Artificial Intelligence Innovation Hub (Artificial Intelligence Institute, Seoul National University)].
	This work was supported by Youlchon Foundation.
	The Institute of Engineering Research at Seoul National University provided research facilities for this work.
	The ICT at Seoul National University provides research facilities for this study.
	U Kang is the corresponding author.
	
	\clearpage
	
	\bibliographystyle{named}
	\bibliography{main}
	
\end{document}